# IMPROVED STATISTICAL BENCHMARKING OF DIGITAL PATHOLOGY MODELS USING PAIRWISE FRAMES EVALUATION

Ylaine Gerardin, John Shamshoian, Judy Shen, Nhat Le, Jamie Prezioso, John Abel, Isaac Finberg, Daniel Borders, Raymond Biju, Michael Nercessian, Vaed Prasad, Joseph Lee, Spencer Wyman, Sid Gupta, Abigail Emerson, Bahar Rahsepar, Darpan Sanghavi, Ryan Leung, Limin Yu, Archit Khosla, Amaro Taylor-Weiner

## EXECUTIVE SUMMARY

Nested pairwise frames is a method for relative benchmarking of cell or tissue digital pathology models against manual pathologist annotations on a set of sampled patches. At a high level, the method compares agreement between a candidate model and pathologist annotations with agreement among pathologists' annotations. This evaluation framework addresses fundamental issues of data size and annotator variability in using manual pathologist annotations as a source of ground truth for model validation. We implemented nested pairwise frames evaluation for tissue classification, cell classification, and cell count prediction tasks and show results for cell and tissue models deployed on a H&E-stained melanoma dataset.

## INTRODUCTION

Digital pathology products built on deep learning algorithms provide a means of quantifying tissue at an unprecedented scale, but acceptance of these products in practice requires rigorous validation. Annotations collected from qualified pathologists are used as ground truth in training and evaluating digital pathology products and their component models. However, the scale of whole-slide histology images, the complexity of model predictions, and the frequency of discordance across individual annotators present challenges to benchmarking algorithms against human performance on the same task.

Scanned images of stained tissue sections are often at a gigapixel scale needed to reveal diagnostic phenotypes at a cellular level. Digital pathology models yield detailed predictions of the tissue microenvironment across entire slides (Figure 1). Because such exhaustive manual annotations of whole-slide images are infeasible to perform, models can instead be evaluated on a representative set of sub-regions, which we term 'frames', sampled across a group of slides[1]. Each frame should be small enough such that exhaustive manual annotations within a frame can be collected and thus compared against model predictions on an equal basis.

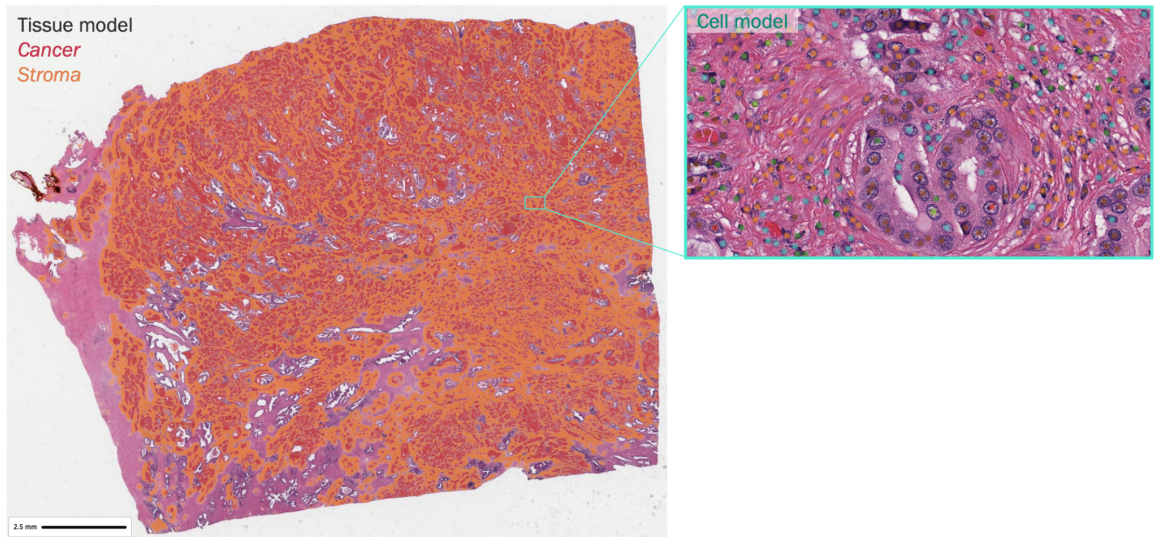

**Figure 1.** An example histopathology image from a prostate H&E slide[2] showing the complexity of tissue (left) and cell (inset, right) model outputs. Predicted substances are shown as colored overlays. An equivalent task of manually annotating every tissue region and cell class in this slide would take a pathologist months to complete.

Because pathologists can disagree on the identity of substances present on a slide, panels consisting of multiple pathologists are used to represent a diversity of opinions and thus reduce bias. The consensus opinion taken across a panel is often treated as a single ground truth, but relying on this consensus can be problematic, for two reasons. First, if a model under evaluation is to be benchmarked against pathologist performance on the same task, the pathologists used in benchmarking need to be independent of the panel used for ground truth determination in order to form a fair comparison to the model. Secondly, rules must be defined to resolve cases where the consensus is ambiguous (e.g., no clear majority in a classification task), and evaluation results may be sensitive to these definitions.

Here we describe a consensus-free framework for relative benchmarking of cell or tissue digital pathology models against manual pathologist annotations on a set of frames. This nested pairwise method calculates an average model performance against each pathologist in the benchmark panel based on pairwise comparisons to every other pathologist. We implemented nested pairwise frames evaluation for tissue classification, cell classification, and cell count prediction tasks and applied this framework to cell and tissue models deployed on an H&E-stained melanoma dataset[2].

## METHODOLOGY

### ANALYSIS PIPELINE

Workflows for the three implemented prediction tasks are shown in Figure 2. Each task consists of one or more preparation steps to yield frame-level pairwise comparisons, followed by application of the nested pairwise statistical framework. The tissue workflow (Figure 2A) can be applied more generally to pixel-level classification tasks. The cell classification workflow (Figure 2A) requires first aligning locations of predicted or annotated cells and can be applied more generally to object classification tasks. The counts workflow (Figure 2C) can generalize to frame-level quantitative features or scores, such as tissue areas. Details on each workflow are provided in the following sections.

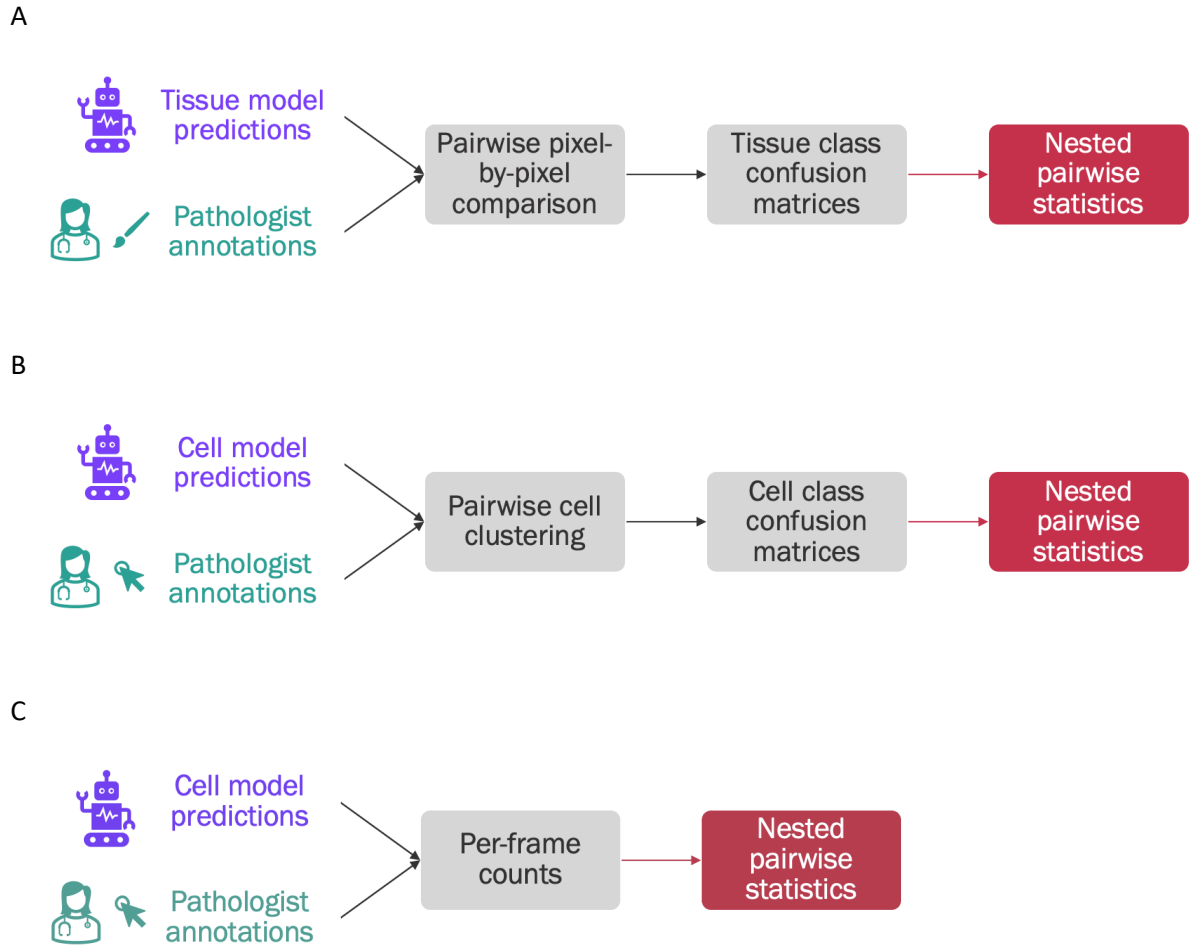

**Figure 2**. Nested pairwise workflows for (**A**) tissue classification, (**B**) cell classification, and (**C**) cell count model evaluation tasks.

## CLASSIFICATION FRAMES

### TISSUE CLASSIFICATION PREPARATION

In tissue frames, annotators have exhaustively identified all areas corresponding to specified tissue types, and the model has assigned each pixel to a single class. Unannotated or unclassified pixels are treated as a background class. Within each frame, a pixel-wise comparison of class assignments is made for each model-pathologist or pathologist-pathologist pair, yielding a set of confusion matrices for each pair. The rows and columns of the confusion matrices correspond to tissue classes assigned by each member of the pair. Each entry in a tissue classification confusion matrix is the number of pixels within the frame corresponding to the indexed classes; entries on the diagonal represent concordance within the pair and off-diagonal entries represent discordance. Confusion matrices are aggregated across frames in common for each pair and used to compute classification metrics in the nested pairwise statistics computation step.

### CELL CLASSIFICATION PREPARATION

In cell frames, the model and the annotators have exhaustively identified and classified all specified cell types present within the frame. For each model-pathologist or pathologist-pathologist pair, cell locations within a frame must be aligned for classifications to be compared. We implemented an iterative pairwise clustering algorithm to perform this alignment. Briefly, we 1) compute pairwise Euclidean distances of all inter-annotators annotations, 2) cluster the pair of annotations with the smallest distance, 3) eliminate the grouped annotations, and 4) repeat steps 2 and 3 until all the pairs of annotations have run out or a pre-specified distance threshold is reached. Figure 3 shows the results of this pairwise alignment procedure for an example frame. Unpaired singleton cells are assigned a 'background' class for the other annotator.

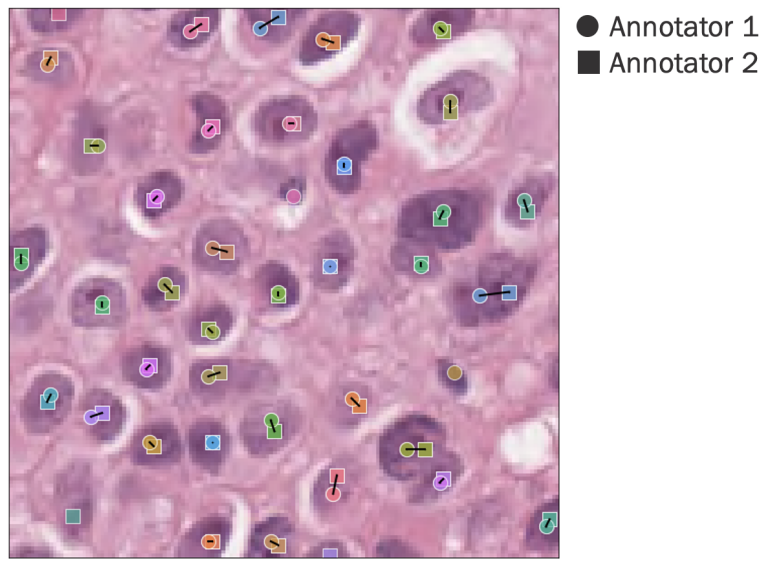

**Figure 3.** Pairwise alignment of annotated cells in an example 150x150 pixel frame. Markers are colored based on assigned clusters. Line segments connect paired objects.

Cell classification confusion matrices are prepared in a similar manner as in the tissue classification task, except that each entry in the matrices corresponds to the number of cells of the corresponding indexed cell classes. These confusion matrices are aggregated across frames in common for each model-pathologist or pathologist-pathologist pair, then used to compute nested pairwise statistics.

### CELL COUNT FRAMES

In the cell counts task, the model and each pathologist assigns a single value per class - the number of cells of that class - to each frame. This matrix of class-wise cell counts per frame across annotators is then used in the computation of nested pairwise statistics.

*NESTED PAIRWISE STATISTICS*

Deriving relative performance metrics using the nested pairwise framework involves outer and inner loop computations (Figure 4). In the outer loop, each pathologist in turn is compared against the model in reference to the set of other pathologists. Each outer loop comparison - model-reference or pathologist-reference - is in turn the average of model-pathologist or comparator-pathologist comparisons, respectively, for each pathologist in the reference set. The overall difference in performance between model and pathologists is the average across the outer loop.

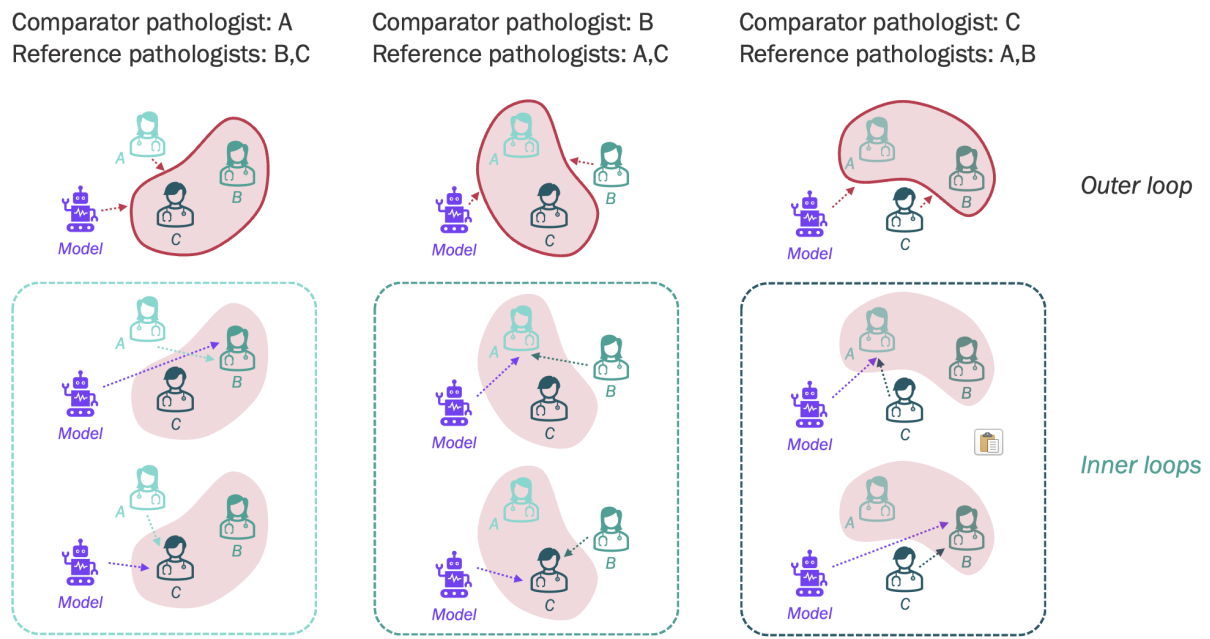

**Figure 4**. The nested pairwise benchmarking procedure shown for a set of three pathologists.

The inner loop comprises pairwise performance metric calculations, considering each pathologist in the reference panel in turn as the ground truth. For classification tasks, we implemented precision, recall, and F1 score on a per-class basis, deriving these metrics from the confusion matrices calculated in the preparation steps. We implemented the intraclass correlation coefficient ICC(2,1) as a metric for pairwise cell count performance.

The nested pairwise method requires at least two pathologists to have annotated each frame, but does not require all pathologists to annotate every frame. For each outer loop computation, both the model and the comparator pathologist are assessed using the same set of frames for an apples-to-apples comparison; these are the frames annotated by the comparator pathologist, as the model is assumed to have been deployed on all images from which frames have been sampled. Both the inner and outer loop averages are weighted by the number of frames used in each computation. Thus, the method accounts for situations where each pathologist in the panel has annotated different subsets of the frames.

We implemented bootstrapping to quantify uncertainty in the nested pairwise statistics, with user-configurable resampling strategies. The default strategy accounts for the hierarchical structuring of slides containing multiple frames by first bootstrapping the slides, and then bootstrapping the frames within each sampled slide. Using the bootstrapped confidence intervals, non-inferiority, equivalence, or superiority testing can be easily applied by setting margins on the average difference in performance between the model and pathologists.

# RESULTS

We applied the nested pairwise evaluation procedure for benchmarking a set of H&E melanoma models. For evaluation of the tissue model's classification performance, 200 375x375 micron frames were sampled across 72 slides. Classification and per-class count performance of the cell model was evaluated on 315 75x75 micron frames sampled across 83 slides. Each frame was annotated by four individual pathologists.

## *TISSUE MODEL EVALUATION*

Results from a single tissue frame are shown in Figure 5, and the summary statistics on the classification metrics are shown in Figure 6. Average differences between the model and the four pathologists across all three performance metrics and all four tissue classes tended to be small (within +/- 0.1, Figure 6B), with the exception of necrosis, where recall was significantly better than pathologists and precision was lower. Higher recall and lower precision implies a tendency of the model to over-call necrosis compared to individual pathologists, which can be seen in the colored overlays in Figure 5.

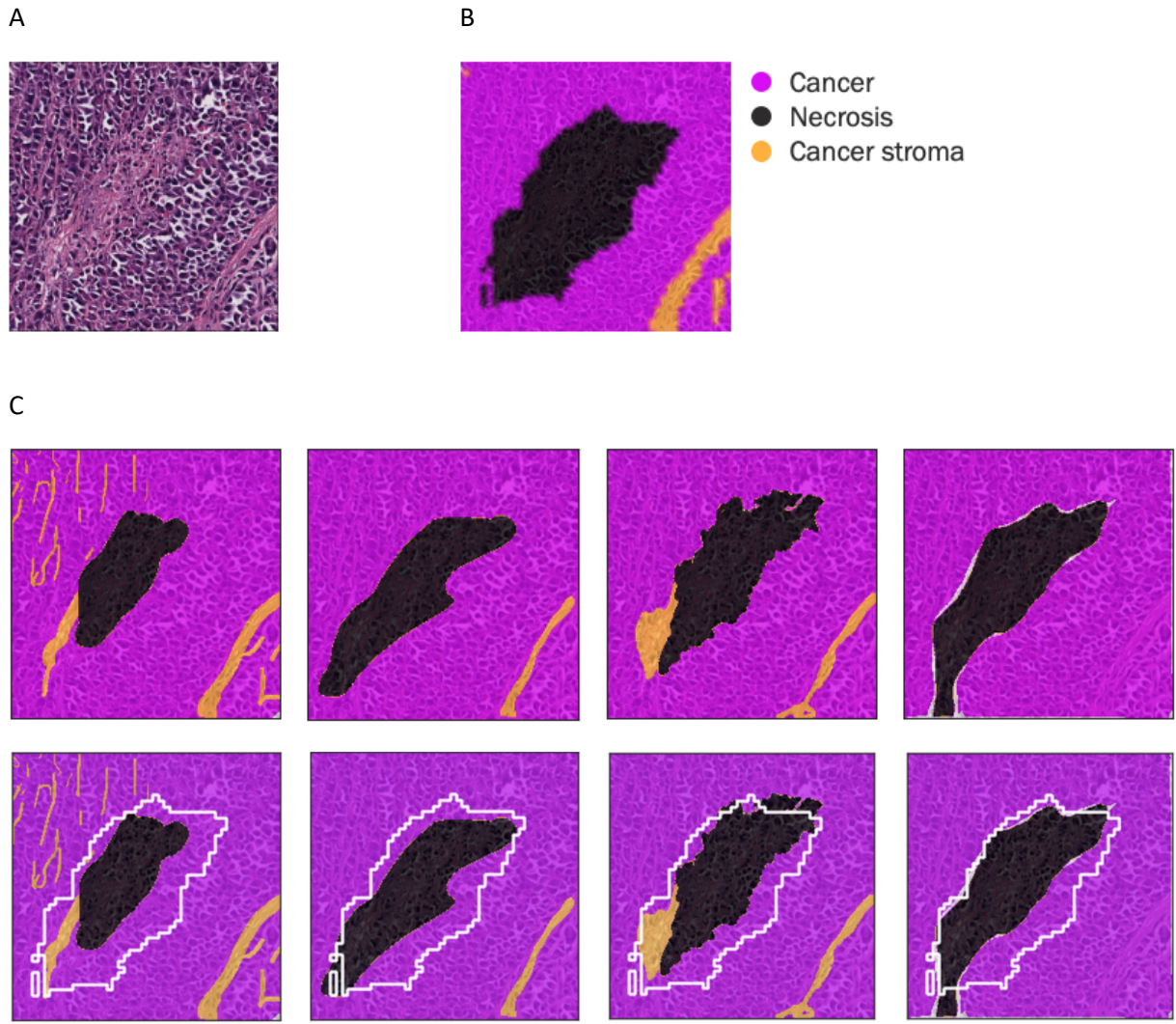

**Figure 5**. Example 375x375 micron melanoma tissue frame: (**A**) original H&E stained image; (**B**) model predictions; (**C**) top row: annotations from four individual pathologists, bottom row: same annotations with model-predicted necrosis region outlined in white.

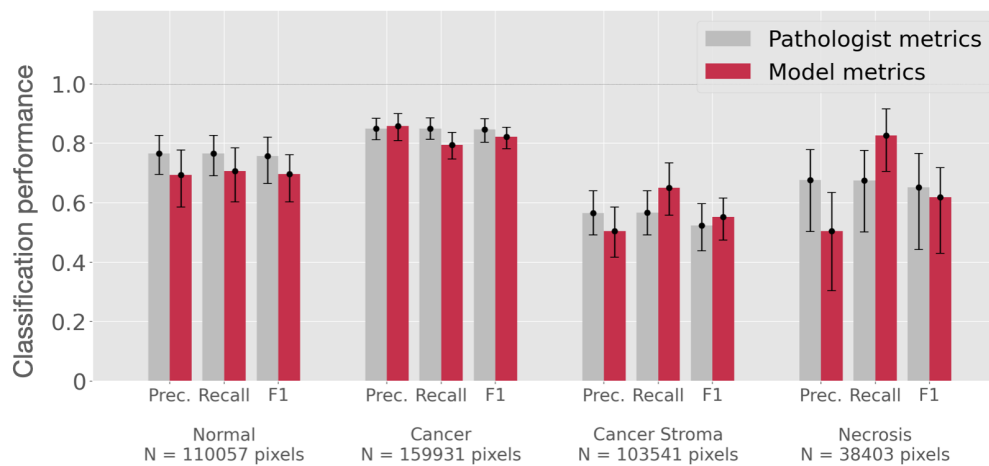

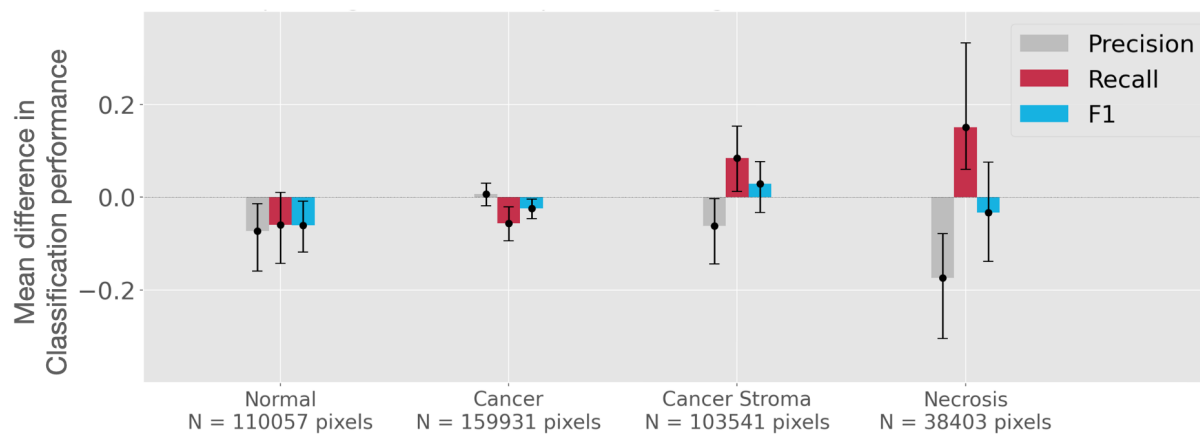

**Figure 6.** Tissue model nested pairwise evaluation results. (**A**) Average performance metrics for model (model) and across pathologists (gray), as assessed by pairwise comparisons to the hold-out reference set. (**B**) Relative performance metrics as assessed by the mean model-pathologist difference in classification metrics relative to the hold-out reference set. Error bars show 95% confidence intervals from bootstrapping across frames. The number of pixels of each tissue type as predicted by the model is indicated below the class names.

### CELL MODEL EVALUATION

Results from evaluating the cell model for classification and cell counts performance across five cell classes are shown in Figure 7. Except for plasma cells, average differences in performance between the model and pathologists tended to be small. Significantly worse precision for model predictions of plasma cells compared to pathologist-pathologist precision suggests over-calling by the model and the potential for further development to improve model performance. Wide confidence intervals on the difference in ICC indicates high variability of model-pathologist agreement in cell counts across frames.

A

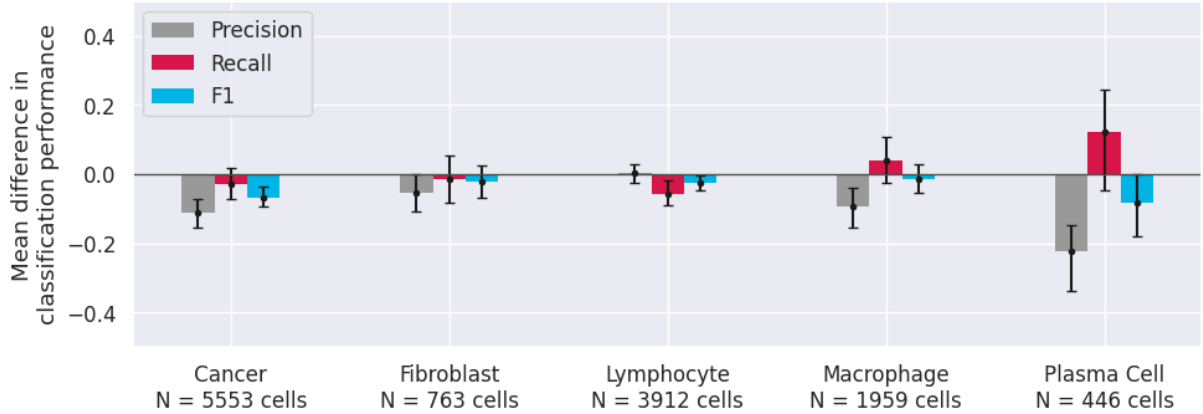

B

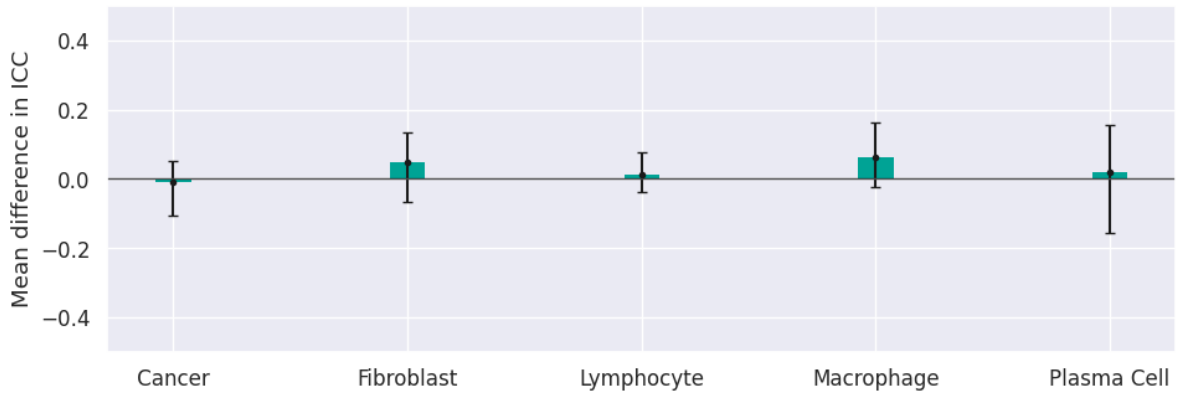

**Figure 7.** Cell model nested pairwise evaluation results for (**A**) classification task as assessed by pairwise precision, recall, and F1 score, along with number of model-predicted cells of each class, and (**B**) count task as assessed by pairwise ICC(2,1) across 315 frames. Plots show the mean model-pathologist difference in performance relative to the hold-out reference set. Error bars show 95% confidence intervals from bootstrapping across frames.

## CONCLUSIONS

As investigations in drug development move beyond standard biomarkers and incorporate machine learning tools, improvements in algorithm evaluation are required to deal with the unique challenges of digital pathology. Nested pairwise frames allow for relative benchmarking of model predictions against annotations collected from multiple pathologists without the need to resolve conflicting annotations into a single consensus. This approach is generalizable to other within-slide classification, object detection, or quantification tasks as well as to slide-level scoring. The resulting statistics can be easily applied towards non-inferiority testing as part of analytical validation, which is an important step towards adoption of digital pathology products.

## REFERENCES

[1] Pokkalla, H. V. et al. (2022). *Systems and methods for frame-based validation*. US 11,527,319 B1

[2] https://www.cancer.gov/tcga